\documentclass[11pt,a4paper]{article}
\usepackage[hyperref]{eacl2021}
\usepackage{times}
\usepackage{latexsym}

\usepackage{microtype}

\usepackage{rotating}
\usepackage{adjustbox}
\usepackage{amsfonts}
\usepackage{multirow} 
\usepackage{tikz}
\usetikzlibrary{calc,shapes,arrows,shadows,shapes.callouts,shapes.arrows,chains,positioning,trees}

\aclfinalcopy 

\title{Better Neural Machine Translation by\\Extracting Linguistic Information from BERT}

\author{Hassan S. Shavarani ~~~~~~~~ Anoop Sarkar\\
  School of Computing Science \\
  Simon Fraser University \\
  BC, Canada \\
  \texttt{\{sshavara,anoop\}@sfu.ca} \\
  }

\date{}

\begin{document}
\maketitle
\begin{abstract}
Adding linguistic information (syntax or semantics) to neural machine translation (NMT) has mostly focused on using point estimates from pre-trained models. Directly using the capacity of massive pre-trained contextual word embedding models such as BERT~\citep{N19-1423} has been marginally useful in NMT because effective fine-tuning is difficult to obtain for NMT without making training brittle and unreliable. We augment NMT by extracting dense fine-tuned vector-based linguistic information from BERT instead of using point estimates. Experimental results show that our method of incorporating linguistic information helps NMT to generalize better in a variety of training contexts and is no more difficult to train than conventional Transformer-based NMT.
\end{abstract}

\section{Introduction}

Probing studies into large contextual word embeddings such as BERT~\citep{N19-1423} have shown that these deep multi-layer models essentially reconstruct the traditional NLP pipeline capturing syntax and semantics \citep{P19-1356};  information such as part-of-speech tags, constituents, dependencies, semantic roles, co-reference resolution information \citep{P19-1452, ICLR2019:EdgeProb} and subject-verb agreement information can be reconstructed from BERT embeddings \citep{Goldberg:BERT}. In this work, we wish to extract the relevant pieces of linguistic information related to various levels of syntax from BERT in the form of dense vectors and then use these vectors as linguistic ``experts'' that neural machine translation (NMT) models can consult during translation.

But can syntax help improving NMT? \citet{Q16-1037,P18-1132, SyntaxInfusedNMT} have reported that learning grammatical structure of sentences can lead to higher levels of performance in NLP models. In particular, \citet{W16-2209} show that augmenting NMT models with explicit linguistic annotations improves translation quality. 

BERT embeddings have been previously considered for improving NMT models.
\citet{D19-5611} replace the encoder token embedding layer in a Transformer NMT model with BERT contextual embeddings. They also experiment with initializing all the encoder layers of the translation model with BERT parameters, in which case they report results on both freezing and fine-tuning the encoder parameters during training. In their experiments BERT embeddings can help with noisy inputs to the NMT model, but otherwise do not help improving NMT performance.

\citet{D19-5603} suggest that replacing the encoder layer with BERT embeddings and fine-tuning BERT while training the decoder leads to a \textit{catastrophic forgetting} phenomenon where useful information in BERT is lost due to the magnitude and number of updates necessary for training the translation decoder and fine-tuning BERT. They present a two-step optimization regime in which the first step freezes the BERT parameters and trains only the decoder while the next step fine-tunes the encoder (BERT) and the decoder at the same time.
\citet{AAAI:BertNMT20} also try to address the \textit{catastrophic forgetting} phenomenon by thinking of BERT as a teacher for the encoder of the neural translation model (student network) \cite{KnowledgeDistillation}. They propose a dynamic switching gate implemented as a linear combination of the encoded embeddings from BERT and the encoder of NMT. However these papers do not really focus on the linguistic information in BERT, but rather try to combine pre-trained BERT and NMT encoder representations.

\citet{SyntaxInfusedNMT} identify \textit{part-of-speech}, \textit{case}, and \textit{sub-word position} as essential linguistic information to improve the quality of both BERT and the neural translation model. They extract each linguistic feature using the Viterbi output of separate models, embed the extracted linguistic information (similar to trained word embeddings) and append these vectors to the token embeddings. However, their model uses point estimates of the syntactic models and they do not use the linguistic information in BERT embeddings.

\citet{AAAI:APT} use multiple multi-layer perceptron (MLP) modules to combine the information from different layers of BERT into the translation model. To make the most out of the fused information, they also alter the translation model training objective to contain auxiliary knowledge distillation \citep{KnowledgeDistillation} parts concerned with the information coming from the pre-trained language model.
\citet{ICLR2020:BERT} also inject BERT into all layers of the translation model rather than only input embeddings. Their model uses an attention module to dynamically control how each layer interacts with the representations.
In both of these works, the training of the Transformer for NMT becomes quite brittle and is prone to diverge to local optima.

In this paper, we propose using pre-trained BERT as a source of linguistic information rather than a source of frozen pre-trained contextual embedding. We identify components of the BERT embeddings that correspond to different types of linguistic information such as part-of-speech, etc. and fine-tune dense vector embeddings for these linguistic aspects of the input and use them within an NMT model. Our approach does not radically complicate the Transformer NMT model training process both in terms of time and hardware requirements and also in terms of training difficulty (avoids bad local optima).

Our contributions are as follows: (1) A method of linguistic information extraction from BERT which needs supervision while training but works without supervision afterwards. (2) An easily trainable procedure for integrating the extracted information into the translation model. (3) Evaluation of the proposed model on small, medium and large translation datasets.

The source code and trained aspect extractors are available at \href{https://github.com/sfu-natlang/SFUTranslate}{https://github.com/sfu-natlang/SFUTranslate} and our experiments can be replicated using scripts under \texttt{resources/ exp-scripts/aspect\_exps}.

\section{NMT and BERT}\label{sec:nmt_and_bert}
Machine translation is the problem of transforming an input utterance sequence \textit{X} in source language $l_f$ into another utterance sequence \textit{Y} (possibly with varying length) in target language $l_e$. Machine translation models search among all possible sequences in target language to find the most probable sequence based on the probability distribution of Equation \ref{eq:nmt}.
\vspace{-0.24cm}
\begin{equation}\label{eq:nmt}
    P(y|X, y\in l_e) = \prod_{i=0}^{|max\ len|} p(y_i|X, y_0, ..., y_{i-1}) 
\end{equation}

Neural machine translation (NMT) tries to model the probability distribution $p(y|X)$ using neural networks by taking advantage of deep learning techniques. Transformers \citep{Transformer} are one type of encoder-decoder neural networks used for translation tasks. In Transformers, the input (in one-hot format) is passed through $N$ layers of encoder and $N$ layers of decoder. In each layer, the layer input passes through multiple attention heads ($h$ heads; each considered a specialist in a different sentence-level linguistic attribute) and then gets transformed to the input for the next layer using a two layer feed-forward perceptron module with input size of $d_{\textit{model}}$ and hidden layer size of $d_{\textit{ff}}$. The final probability distribution $p(y|X)$ is generated using an affine transformation applied to the output of the last feed-forward module in the $N^{th}$ decoder layer. Please see \citep{Transformer} for further details.

BERT \citep{N19-1423} adopts the encoder part of the transformer model and requires training it on large amounts of text data using a \textit{masked language model} objective over sub-words $p(y_i|X, y_0, ..., y_{i-1}, y_{i+1}, ..., y_{max\ len})$ instead of guessing the next sub-word $p(y_i|X, y_0, ..., y_{i-1})$. This bidirectional context turns BERT into a provider of strong contextual sub-word embeddings in many languages.
These massively over-parameterized neural networks have revolutionized many different NLP tasks. Effective application of BERT in NMT has been studied in a number of contemporary research projects; Language Modeling, Named Entity Recognition, Question Answering, Natural Language Inference, Text Classification \cite{N19-1423}, and Question Generation \cite{W19-8624}. We approach this problem from the novel perspective of extracting linguistic information encoded in BERT and applying such information in NMT.

\section{Linguistic Aspect Extraction from BERT}
Since BERT contextual embeddings contain a variety of information (linguistic and non-linguistic), extraction of relevant information plays an important role in further improvement of the downstream tasks. In the rest of this section, we define \textit{aspect vectors} as single-purpose dense vectors of extracted linguistic information from BERT, discuss how aspect vectors can be extracted, and explain how to integrate aspect vectors into NMT.

\subsection{Aspect Vectors}\label{sec:aspect}
To start the information extraction process, we initially need to choose a limited (desired) set of linguistic attributes to look for in BERT embeddings. This attribute set can contain a number of linguistic aspects (e.g. part-of-speech). Each linguistic aspect itself will be defined over a possible aspect tag set (e.g. the set of \{\textit{NOUN}, \textit{ADJ}, ...\} in part-of-speech). In this paper, we show a linguistic attribute set with $\mathbb{A}$, show a generic aspect with $a$ and point to its relative tag set with $t_a$.

Given the definition of a linguistic aspect and inspired by the information bottleneck idea \citep{InformationBottleneck}, we define an \textit{aspect vector} as a single-purpose dense vector extracted from BERT and containing information about a certain linguistic aspect of a particular (sub-word) token in the input sequence. Aspect vectors can be interpreted as feature values equivalent to a specific key (aspect).

\subsection{Aspect Vector Extraction}\label{sec:feat_extract}

For each embedding vector $\mathbf{E}$ and linguistic aspect $a$, we define $M_a$ as an aspect-extraction function where $\mathbf{e}_a=M_a(\mathbf{E})$ is a single-purpose dense vector containing maximum aspect information and minimum irrelevant other information. 

We ensure the aspect encoding power of $\mathbf{e}_a$ by retrieving its equivalent tag in $t_a$ using a classifier. The aspect prediction loss for a linguistic attribute set $\mathbb{A}$ of size $n$ can be calculated as the average cross entropy loss ($\mathcal{L}_{CE}$) between the classifier prediction and the expected aspect tags for each aspect (Equation \ref{eq:pred_loss}).

\begin{equation}\label{eq:pred_loss}
    \mathcal{L}_{a}=\frac{1}{n}\sum_{i=0}^{|n|}\mathcal{L}^{i}_{CE}
\end{equation}

We also ensure information integrity\footnote{We don't expect $M_a$ to change the information inside $\mathbf{E}$ but rather to extract the relevant information.} of $\mathbf{e}_a$ by concatenating all the aspects (in addition to a ``\textit{left-over}'' aspect equivalent to all the other non-interesting information) and reconstructing the original embedding vector $\mathbf{E}$ from them\footnote{This idea is analogous to stack-propagation \citep{P16-1147} in which propagating the information loss for two tasks helps improving the quality of the encoded representations.} in reconstruction vector $\mathbf{R}$. The reconstruction loss ($\mathcal{L}_{r}$) for the extracted aspect vectors can be calculated as the euclidean distance of the reconstruction vector $\mathbf{R}$ and the original embedding vector $\mathbf{E}$ (Equation \ref{eq:reconstruct_loss}).

\begin{figure}
    \centering
    \input{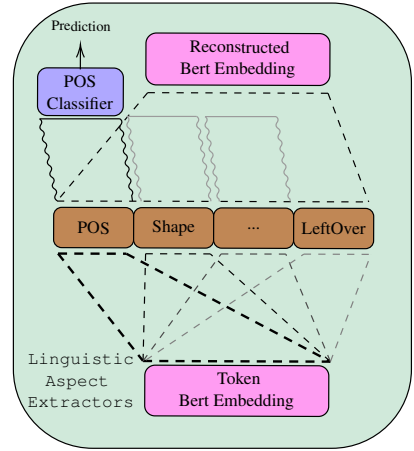}
    \caption{Schematic Aspect Extraction from BERT}
    \label{fig:feat_ext}
    \vspace{-0.5cm}
\end{figure}

\begin{figure*}
    \centering
    \input{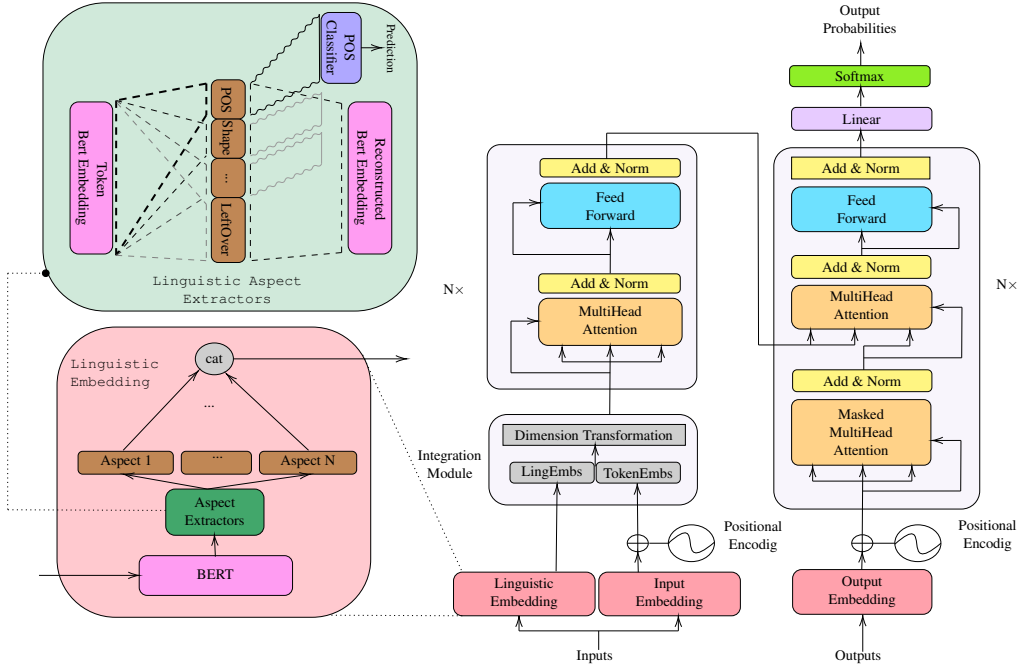}
    \caption{Integration of Extracted Aspect Vectors into NMT. The right hand side part of this figure is taken from \citet{Transformer}.}
    \label{fig:integrate}
    \vspace{-0.3cm}
\end{figure*}

\begin{equation}\label{eq:reconstruct_loss}
    \mathcal{L}_{r}=||\mathbf{R}-\mathbf{E}||^2
\end{equation}

In addition, since our aspect extractor is similar in architecture to a multi-head attention module (with a difference in the fact that we know what exactly each head will be responsible for), to prevent learning redundant representations \citep{NIPS:MHA}, we add the average euclidean similarity  ($\mathcal{L}_{s}$) of each pair of aspect vectors to the training loss function (Equation \ref{eq:dist_loss}).

\begin{equation}\label{eq:dist_loss}
    \mathcal{L}_{s}=1-\left(\frac{1}{n(n-1)}\sum_{i=0}^{|n|}{\sum_{j\neq i=0}^{|n|}{||{e}_i - {e}_j||^2}}\right)
\end{equation}

The aspect extractor will be trained over the accumulation of the three mentioned loss components (Equation \ref{eq:feat_extract_loss}). 
Figure \ref{fig:feat_ext} demonstrates different parts of the aspect extractor and their connections.

\begin{equation}\label{eq:feat_extract_loss}
    \mathcal{L}_{fe} = \mathcal{L}_{a} + \mathcal{L}_{r} + \mathcal{L}_{s}
\end{equation}

As another important point, a pre-trained BERT model has multiple encoder layers as well as an embedding layer. Choosing the proper layer which contains all of our desired aspects is not simply possible since different layers specialize in different linguistic aspects \citep{P19-1356, P19-1452}.

Therefore, as \citet{N18-1202} suggest, we define BERT embedding vector $\mathbf{E}$ as a weighted sum of all BERT layers (of size $\ell$) using Equation \ref{eq:bert} where $\alpha$ weights are learnable parameters and will be trained along with the other aspect extractor parameters.
\begin{equation}\label{eq:bert}
    \mathbf{E} = \sum_{j=0}^{\ell} \alpha_{j} \mathbf{E}^{BERT}_{j}
\end{equation}

\subsection{Integrating Aspect Vectors into NMT}\label{sec:integrate}

Once the aspect vectors are created, we throw away the classifiers and the reconstruction layers and place the encoder part of our trained aspect extractor (the mapping from BERT contextual embeddings to aspect vectors) in an input integration module designed to augment the neural translation model input with aspect vectors\footnote{We use the same sub-word model in pre-trained BERT to provide sub-word tokens to our NMT model.}.

The integration module (constructed using a two layer perceptron network) receives the concatenated aspect vectors (we call this concatenated vector a linguistic embedding\footnote{This embedding vector can be similar to what a factor token contains in Factored-NMT \cite{FNMT} with a difference that it is generated in the space of linguistic aspects and does not need an embedding layer.}) and the token embedding (inherited from the Transformer model), and maps the linguistic embedding into a vector of the same size as the token embedding. Then, it projects the concatenation of both embeddings to a vector with the same size as the token embedding of the original Transformer model\footnote{This step is necessary to prevent any change in other parts of the model which would make comparison of the results unfair due to effects on the number of parameters and the learning capability of the model.}. Figure \ref{fig:integrate} demonstrates this process.

\section{Experiments}
In this section, we initially examine our designed aspect extractor and report its classification accuracy scores. Next, we integrate the extracted aspect vectors into the neural machine translation framework as explained in Section \ref{sec:integrate} and study the effects of integrated vectors on the performance of the models. 

\subsection{Data}\label{sec:data}

We choose three German (which has explicit and nuanced linguistic features) to English datasets in different data sizes to examine our proposed framework.

We use Multi30k (M30k)\footnote{AKA \textit{Flickr30K} provided in task 1 of WMT17 multimodal machine translation, \href{http://www.statmt.org/wmt17/multimodal-task.html}{http://www.statmt.org/wmt17/ multimodal-task.html}} as our small dataset. This dataset contains a multilingual set of image descriptions in German, English and French. Due to this reason, we also consider experimenting on German to French as our second small dataset. The M30k data contains 29K training sentences, 1014 validation sentences ($val$) and 1000 test sentences ($test2016$).

We take IWSLT \citep{IWSLT_data}\footnote{2017 was the last year that the data for this task got updated; \href{https://wit3.fbk.eu/mt.php?release=2017-01-mted-test}{https://wit3.fbk.eu/mt.php?release=2017-01-mted-test}} as our medium sized dataset. The sentences in this dataset are quite different from M30k since they are composed from the transcriptions of TED talks as well as dialogues and lectures\footnote{While the talks are quite polished, they still contain many verbal structures and sometimes even sounds (e.g. ``\texttt{Imagine an engine going clack, clack, clack, clack, clack, clack, clack.}'').}. The IWSLT data contains 208K training sentences, 888 validation sentences ($dev2010$) and multiple test sets ($tst2010$ to $tst2015$ with 1568, 1433, 1700, 993, 1305, and 1080 sentences, respectively).

For the large data size, we consider WMT\footnote{Europarl+CommonCrawl+NewsCommentary \href{https://www.statmt.org/wmt14/translation-task.html}{https://www.statmt.org/wmt14/translation-task.html}, please note that in the later years this training set remained the same, but ParaCrawl data was added to it. We do not use ParaCrawl data since it is quite noisy and we aim to limit the effects of uncontrolled variables in our training data. However, we report our results on all the test tests after 2014.}, a large (4.5M training sentences) set of parallel sentences from the proceedings of the European Parliament as well as web crawled news articles. We remove 0.05\% of the training data (2290 sentences; lines with numbers divisible by 2000) and use it as the validation set (we call it $wmt\_val$) and take $newstest$ data from 2014 to 2019 as our test sets (with 3003, 2169, 2999, 3004, 2998 and 1997 sentences, respectively).

We remove train data sentences longer than 100 words and uncase and normalize both side sentences using \texttt{MosesPunctNormalizer}\footnote{\href{https://github.com/alvations/sacremoses/blob/master/sacremoses/normalize.py\#L11}{https://github.com/alvations/sacremoses/}} before tokenization. The reference side of the test data remains untouched in all the steps of our experiments.

\subsection{Linguistic Aspect Vector Extraction}\label{sec:lave}
In this section, we study our linguistic aspect extractor training procedure and analyze the quality of the extracted aspect vectors.

We choose our linguistic attribute set ($\mathbb{A}$) as \citet{SyntaxInfusedNMT} suggest, however, we replace `\textit{case}' with `\textit{word-shape}'\footnote{Representing capitalization (changing alphabet to \texttt{x} or \texttt{X}), punctuation, and digits (changing digits to \texttt{d}). As an example for \textit{word-shape}, the sub-word \textit{\#\#arxiv.} in the token `\textit{myarxiv.org}' will turn to \textit{\#\#xxxxx.}.} since we believe the complete shape of the word is much more informative specially in sub-word settings. In addition, we consider a two-level hierarchy in part-of-speech tags to benefit from both higher accuracy in exploring the syntactic search space and lower model confusion in cases where the fine-grained tags are not helpful. Therefore, we consider coarse-grained and fine-grained \textit{part-of-speech} (CPOS and FPOS), \textit{word-shape} (WSH), and \textit{sub-word position}\footnote{Encoding the word with one of the three labels ``\texttt{Begin}'', ``\texttt{Inside}'', or ``\texttt{Single}''.} (SWP) to form our experimental linguistic attribute set ($\mathbb{A}$). Other linguistic attributes such as dependency parses or sentiment could be considered as aspects in our model but we leave that for future work.

\begin{table*}
    \centering
    \begin{tabular}{llllllllll}
    \hline\hline
    \multirow{2}{*}{} & \multicolumn{5}{c}{\textbf{Sub-word Level}} & \multicolumn{4}{c}{\textbf{Word Level}} \\\cline{2-10} 
    & \multicolumn{1}{c}{\textbf{CPOS}} & \multicolumn{1}{c}{\textbf{FPOS}} & \multicolumn{1}{c}{\textbf{WSH}} & \multicolumn{1}{c}{\textbf{SWP}}
    & \multicolumn{1}{c}{\textbf{\texttt{\#tokens}}} &
    \multicolumn{1}{c}{\textbf{CPOS}} & \multicolumn{1}{c}{\textbf{FPOS}} & \multicolumn{1}{c}{\textbf{WSH}} & \multicolumn{1}{c}{\textbf{\texttt{\#tokens}}}\\\hline
    {M30k}  & 96.88 & 96.18 & 99.79 & 99.93 & 16096 & 97.95 & 97.34 & 99.74 & 12823 \\
    {IWSLT} & 92.69 & 90.48 & 99.73 & 97.14 & 22687 & 94.84 & 93.07 & 99.69 & 19039\\
    {WMT}   & 92.64 & 91.60 & 97.74 & 98.94 & 70139 & 94.86 & 94.01 & 97.38 & 55135 \\\hline\hline
    \end{tabular}
    \caption{F-1 scores acquired after training the aspect extractor on German side of parallel data and passing the validation sets of each data set through trained aspect extractors. The \texttt{\#tokens} column shows the number of tokens in the validation set.}
    \label{tab:feat_ext_results}
    \vspace{-0.3cm}
\end{table*}

\begin{table}
    \centering
    \begin{tabular}{lll}
    \hline\hline
        \textbf{\begin{tabular}[c]{@{}c@{}}Aspect Extractor\\Training Data\end{tabular}} & \textbf{FPOS} & \textbf{SWP} \\ \hline
        M30k & 79.39 & 90.63 \\
        IWSLT & 77.80 & 88.34 \\
        WMT & 82.13 & 91.42\\
        TIGER & 84.64 & 92.64\\\hline\hline
    \end{tabular}
    \caption{F-1 scores of fine-grained part-of-speech prediction of TIGER corpus test data (BERT encoded) fed to each of the trained aspect classifiers. The scores are calculated over a total of 7516 sub-word tokens in 358 test sentences of TIGER. Extractors trained on M30k, IWSLT, and WMT have not been provided with any part of TIGER before evaluation.}
    \label{tab:TIGER}
    \vspace{-0.3cm}
\end{table}

We use the spaCy German tagger\footnote{\href{https://spacy.io/models/de\#de_core_news_md}{https://spacy.io/models/de}} model to acquire our intended linguistic aspect labels. Since spaCy is trained on word-level while BERT is trained on sub-word level, we had to align the sequences using a monotonic alignment algorithm (see Appendix \ref{sec:apx_aspect_details}). The fine-grained part-of-speech tagger in spaCy\footnote{SpaCy reports 96.52\% accuracy for this model.} is pre-trained on TIGER Corpus\footnote{\href{https://www.ims.uni-stuttgart.de/forschung/ressourcen/korpora/tiger/}{https://www.ims.uni-stuttgart.de/}} \citep{TIGER} and inherits its 55 fine-grained tags from TIGER treebank. The coarse-grained spaCy part-of-speech tagger has been trained by defining a direct mapping from 55 tags of the TIGER treebank to the 16 tags in the Universal Dependencies v2 POS tag set\footnote{\href{https://universaldependencies.org/v2/postags.html}{https://universaldependencies.org/v2/postags.html}}.

We use a 12-layer\footnote{Hidden state size of 768 with 12 heads; written in PyTorch and distributed by \citet{HuggingFacesTS}. You can find model configurations in \href{https://github.com/dbmdz/berts}{https://github.com/dbmdz/berts}.} German pre-trained BERT model for encoding the source sentences in aspect extractors. 
We use an uncased model as our translation model performs on lowercased data and the results are recased using the moses recaser so that the results are cased BLEU scores comparable to other systems\footnote{We recommend using a cased BERT model for translation systems that handle casing differently.}.
We pass the BERT-encoded source sentences through a single perceptron middle layer of size 1000. We divide the output of this layer to `\textit{number of aspects + 1}' splits to form our desired aspect vectors (of size 200). Please see Appendix \ref{sec:apx_aspect_details} for more implementation details.

We train three different aspect extractors, one for each dataset and feed in the source sentences of the dataset to our model in batches of size 32 for 3 epochs\footnote{Since the number of WMT sentences are much bigger, we stop training WMT aspect extractors when there is no improvement in aspect classification result (rounded to have 3 decimal places) of any label for at least 40 batches.}. 
Table \ref{tab:feat_ext_results} shows F-1 scores of classifying the validation set data using different aspect vectors after training the aspect extractors on the train set sentences. Please note that for calculating the word-level scores, in cases of disagreement between different sub-word tokens, the sub-word prediction of the first sub-word token has been counted as the prediction for the word label.

We also validate our trained (on M30k, IWSLT, and WMT) aspect extractors against the manual annotations of TIGER treebank with which the spaCy fine-grained part-of-speech tagger has been trained.
We train an extra aspect extractor using the train set of TIGER corpus and test all four trained aspect extractors against TIGER data test set\footnote{We use \texttt{german\_tiger\_test\_gs.conll} in the version of TIGER released in \textit{2006 CoNLL Shared Task - Ten Languages}. Both train and test data are accessible through \href{https://catalog.ldc.upenn.edu/LDC2015T11}{https://catalog.ldc.upenn.edu/LDC2015T11}.}. This experiment evaluates the absolute power of our simple feed-forward aspect extractors in performing the aspect classification task. Please note that our goal in this experiment is not to achieve the state-of-the-art fine-grained part-of-speech tagging results as our aspect extractors receive their input from BERT and do not directly access the tagged input sentences. Table \ref{tab:TIGER} contains the results of comparison between predictions of different aspect extractor classifiers and TIGER gold labels.

\subsection{Uniqueness of Information in Linguistic Aspect Vectors}

Considering the high F-scores for each aspect category in each dataset (Table \ref{tab:feat_ext_results}), we can conclude that our aspect extractor maximizes the relevant information extraction from BERT embeddings. The loss in Equation \ref{eq:dist_loss} maximizes the distance between aspect vectors. To test whether this leads to a diverse set of aspect vectors, each specialized to their own linguistic attributes, we consider each aspect category $a$, after training the aspect extractors. We take each of the other extracted aspect vectors $a'$ (except the ``\textit{left-over}'' vector) and use each of them to train a new classifier\footnote{We thank the anonymous reviewers for their valuable feedback on this procedure.} that predicts the right class for category $a$ based on aspect vector $a'$. This will test the correlation between the information in aspect vectors $a'$ and the tags in category $a$. If the classification scores for this counterfactual test are high then our model has failed in fine-tuning each aspect vector to predict a particular linguistic aspect. We compare the classification scores to a trivial baseline: predict the most frequent class always. Table \ref{tab:sanity_check} shows the results of this counterfactual test on the aspect extractor trained on TIGER data. We can see that the average F-1 scores are very low when we use counterfactual aspect vectors to predict a linguistic aspect on which it was not fine-tuned (e.g. use aspect vector trained on part-of-speech to predict word shape). This shows that our training method fine-tunes each aspect vector to its linguistic task.

\begin{table}
    \centering
    \addtolength{\tabcolsep}{-2pt} 
        \begin{tabular}{ccccc}
        \hline\hline
            \multicolumn{1}{c}{\multirow{2}{*}{\textbf{TIGER test}}} & \multicolumn{4}{c}{\textbf{Sub-word Level}} \\ \cline{2-5} 
            \multicolumn{1}{c}{} & \textbf{CPOS} & \textbf{FPOS} & \textbf{WSH} & \textbf{SWP} \\ \hline
            \begin{tabular}[c]{@{}c@{}}most frequent\\class\end{tabular} & NOUN & NN & xxxx & single\\
            \begin{tabular}[c]{@{}c@{}}percentage\\in total\end{tabular} & 27.12 & 27.07 & 39.07 & 59.92\\\hline
            \begin{tabular}[c]{@{}c@{}}average\\classification F-1\end{tabular} & 1.89 & 0.23 & 12.20 & 42.97\\
            \texttt{\#tokens} &  \multicolumn{4}{c}{7516 $\times$ 3 = 22548}\\ \hline\hline
            
        \end{tabular}
    \caption{Classification scores of each aspect classifier when fed with other extracted aspect vectors. We expect the F-1 scores to be low so we can conclude that our aspect extractor truly excludes irrelevant information from each aspect.}
    \label{tab:sanity_check}
\end{table}

To validate the loss in Equation \ref{eq:reconstruct_loss}, we calculate the average euclidean distance of the aspect extractor reconstructed vectors and the original BERT embedding vectors\footnote{Average results of Equation \ref{eq:reconstruct_loss} for all the tokens in the train set.} for M30k German to English dataset. We unit normalize each of the vectors for a score in $[0,1]$. The average euclidean distance value of 0.1863 tells us that the reconstruction component of the aspect extractor is capable of reconstructing vectors that are close to the original embedding vectors.

\subsection{Linguistic Aspect Integrated Machine Translation}\label{sec:laimt}

\begin{table*}[ht]
\centering
\small{
    \addtolength{\tabcolsep}{-3pt}
    \renewcommand{\arraystretch}{1.2}
    \begin{tabular}{lccccccccc}
    \hline\hline
    \multirow{2}{*}{\textbf{a) M30k\textsuperscript{$\dagger$}}}& \multicolumn{4}{c}{\textbf{German to English}} & &\multicolumn{4}{c}{\textbf{German to French}} \\ \cline{2-5} \cline{7-10} 
    & \textbf{val} & \textbf{test2016} & \textbf{\texttt{\#param}} & \textbf{\texttt{runtime}}\textsuperscript{$*$} & & \textbf{val} & \textbf{test2016} & \textbf{\texttt{\#param}} & \textbf{\texttt{runtime}}\textsuperscript{$*$} \\ \hline
    \begin{tabular}[c]{@{}l@{}}\citealt{Transformer}\end{tabular}                         & 39.63 & 38.35 & ~~9.5 M  & ~~84 min & &  31.07 & 30.29 & ~~9.4 M  & ~~93 min \\
    \begin{tabular}[c]{@{}l@{}}\citealt{SyntaxInfusedNMT}\end{tabular}                    & 40.03 & 38.32 &  13.9 M  & 514 min  & &  32.55 & 32.71 &  13.6 M  &  504 min \\
    \begin{tabular}[c]{@{}l@{}}\citealt{D19-5611}\\\ \ (bert freeze)\end{tabular}         & 40.07 & 39.73 & ~~9.1 M  & ~~99 min & &  33.83 & 33.15 & ~~9.0 M  &  104 min \\ 
    \begin{tabular}[c]{@{}l@{}}Aspect Augmented \\\ \ \  +M30k asp.\ vectors\end{tabular} & \textbf{40.47} & 40.19 &  10.1 M  & 104 min  & &  34.45 & \textbf{34.42} & ~~9.9 M  &  108 min \\ 
    \begin{tabular}[c]{@{}l@{}}Aspect Augmented \\\ \ \  +WMT asp.\ vectors\end{tabular}  & 38.72 & \textbf{41.53} &  10.1 M  & 102 min  & &  \textbf{34.73} & 34.28 & ~~9.9 M  &  118 min \\ 
    \hline\hline
    ~\\
    \hline\hline
    \textbf{b) IWSLT\textsuperscript{$\dagger$}}& \textbf{dev2010} & \textbf{tst2010} & \textbf{tst2011} & \textbf{tst2012} & \textbf{tst2013} & \textbf{tst2014} & \textbf{tst2015} & \textbf{\texttt{\#param}} & \textbf{\texttt{runtime}}\textsuperscript{$*$} \\ \hline
    \begin{tabular}[c]{@{}l@{}}\citealt{Transformer}\end{tabular}                          & 27.69 & 27.93 & 31.88 & 28.15 & 29.59 & 25.66 & 26.76 & 18.4 M & ~~172 min \\
    \begin{tabular}[c]{@{}l@{}}\citealt{SyntaxInfusedNMT}\end{tabular}                     & 29.53 & 29.67 & 33.11 & 29.42 & 30.89 & 27.09 & 27.78 & 28.9 M &  1418 min \\
    \begin{tabular}[c]{@{}l@{}}\citealt{D19-5611}\\\ \ (bert freeze)\end{tabular}          & 30.31 & 30.00 & 34.20 & 30.04 & 31.26 & 27.50 & 27.88 & 18.0 M & ~~212 min \\
    \begin{tabular}[c]{@{}l@{}}Aspect Augmented \\\ \ \  +IWSLT asp.\ vectors\end{tabular} & 29.03 & 29.17 & 33.42 & 29.58 & 30.63 & 26.86 & 27.83 & 18.9 M & ~~214 min \\
    \begin{tabular}[c]{@{}l@{}}Aspect Augmented \\\ \ \  +WMT asp.\ vectors\end{tabular}   & \textbf{31.22} & \textbf{30.82} & \textbf{34.79} & \textbf{30.29} & \textbf{32.34} & \textbf{27.71} & \textbf{28.40} & 18.9 M & ~~211 min \\
    \hline\hline
    ~\\
    \hline\hline
    \textbf{c) WMT\textsuperscript{$\dagger$}}& \textbf{wmt\_val} & \textbf{nt2014} & \textbf{nt2015} & \textbf{nt2016} & \textbf{nt2017} & \textbf{nt2018} & \textbf{nt2019} & \textbf{\texttt{\#param}} & \textbf{\texttt{runtime}}\textsuperscript{$*$} \\ \hline
    \begin{tabular}[c]{@{}l@{}}\citealt{Transformer}\end{tabular}                          & 28.96 & 26.91 & 26.93 & 31.42 & 28.07 & 33.56 & 29.77 & 68.7 M & ~~35 h \\ 
    \begin{tabular}[c]{@{}l@{}}\citealt{SyntaxInfusedNMT}\end{tabular}                     & 28.56 & 27.80 & 26.93 & 30.44 & 28.63 & 33.87 & 30.48 & 93.8 M &  258 h \\
    \begin{tabular}[c]{@{}l@{}}\citealt{D19-5611}\\\ \ (bert freeze)\end{tabular}          & 28.63 & 27.54 & 27.15 & 31.69 & 28.30 & 33.89 & \textbf{31.48} & 69.1 M & ~~33 h \\ 
    \begin{tabular}[c]{@{}l@{}}Aspect Augmented \\\ \ \  +WMT asp.\ vectors\end{tabular}   & \textbf{28.98} & \textbf{28.05} & \textbf{27.58} & \textbf{32.29} & \textbf{29.07} & \textbf{34.74} & \textbf{31.48} & 70.3 M & ~~46 h \\ 
    \hline\hline    
    \end{tabular}
    
    \caption{Evaluated cased \textsc{Bleu} score (calculated using \texttt{mteval-v14.pl} script) results on M30k, IWSLT, and WMT datasets. \texttt{\#param} represents the number of trainable parameters (size of BERT model parameters [110.5M] has not been added to the model size for the aspect augmented and bert-freeze models since BERT is not trained in these settings). \texttt{runtime} is the total time the training script has ran and includes time taken for reading the data and training the model from scratch (iterating over the instances for all the epochs).\\\textit{All the baseline results are achieved using our re-implementation of the mentioned papers.}\\\textsuperscript{$*$} We have used a single GeForce GTX 1080 GPU for M30k experiments and a single Titan RTX GPU for IWSLT and WMT experiments.\\\textsuperscript{$\dagger$} Each experiment was repeated three times, and we report the average in this table.\label{tab:results}}
    }
    \vspace{-0.3cm}
\end{table*}

After confirming the adequacy and uniqueness of linguistic information in aspect vectors, we integrate the encoder part of aspect extractors into the translation model and perform translation experiments on M30k, IWSLT, and WMT datasets. In our experiments, we compare our model to three baselines : (1) the vanilla transformer model \citep{Transformer} which does not use any external source of information, (2) the syntax-infused transformer model \citep{SyntaxInfusedNMT} which explicitly embeds linguistic aspect labels and concatenates their embedding to the token embedding, (3) the transformer model with bert-freeze input setting \citep{D19-5611} which replaces the input embedding layer of the encoder module in transformer with a fully pre-trained BERT model. Appendix \ref{sec:apx_nmt_details} provides the configurations and sufficient details for replication of our experiments in this section.

During each training trial, we perform 9 validation set evaluation steps (one after visiting each 10\% of the data). In each step, the validation set is translated with the current state of the model (at the time of evaluation) and the generated sentences are detokenized and compared to the validation set reference data to produce sentence-level \textsc{Bleu} \citep{Bleup1} scores. The best scoring model throughout training is selected as the model with which the test set(s) are translated.

For M30k and IWSLT data sets, we train two separate models, one using the aspect vectors trained on the source side of its own training data (in-domain) and the other using the aspect vectors trained on the source side of WMT data (out-of-domain). We use cased \textsc{Bleu} (evaluated with the standard \texttt{mteval-v14.pl} script) and \textsc{METEOR} \cite{METEOR} to compare different models. Tables \ref{tab:results} and \ref{tab:meteor_results} show the results of evaluating the models trained with different mentioned settings. 

The evaluation results show that taking advantage of aspect vectors improves the accuracy of translating German to both English and French in M30k as well as German to English in IWSLT and WMT. Also, in majority of the cases WMT-trained aspect vectors have pushed the model to produce more accurate results since they contain more generalized information. Based on these results, we conjecture that aspect vectors trained on large out-of-domain data can be helpful in low-resource settings but we leave the examination of this idea for future work.

\begin{table*}[th]
  \centering
  \small{
  \begin{tabular}{rl}
    \hline\hline 
    Source & Ihm \textbf{werde weiterhin vorgeworfen}, unerlaubt geheime Informationen weitergegeben zu haben.\\ 
    Reference& He \textbf{is still accused of passing on} secret information without authorisation.\\\hline
    \citealt{Transformer} & He has \textbf{also been accused of having} illegally \textbf{passed on} secret information.\\
    \citealt{D19-5611} & He \textbf{continues to be accused of} fraudulently \textbf{passing on} secret information.\\
    \citealt{SyntaxInfusedNMT} & He \textbf{is also accused of having pass} unauthorised secret information \textbf{on}.  \\
    Aspect Augmented NMT& He \textbf{is still accused of passing on} illegal secret information.\\
    \hline\hline
    Source & Auto und Traktor krachen zusammen: Frau stirbt bei schrecklichem Unfall\\
    Reference & Car and tractor \textbf{crash together}: woman \textbf{dies} in terrible accident\\\hline 
    \citealt{Transformer} & Car and \underline{traktor} \textbf{cranes together}: women \textbf{die} in the event of a terrible accident.\\
    \citealt{D19-5611} & Cars and tractors \textbf{are killing} women in the event of a terrible accident.\\
    \citealt{SyntaxInfusedNMT} & Auto and tractor \textbf{are blowing together}: woman \textbf{dies} when the terrible accident occurs.\\
    Aspect Augmented NMT& Car and tractor \textbf{crash together}: woman \textbf{dies} in terrible accidents.\\
    \hline\hline

  \end{tabular}
  }
  \caption{Examples of improved translation quality of WMT data where \textit{part-of-speech} aspect vectors have helped the model choose better words both syntactically and semantically.}
  \label{table:example1}
\end{table*} 

\begin{table*}[th]
  \centering
  \small{
  \begin{tabular}{rl}
    \hline\hline
    Source & Bucht die besten Hostels in \textbf{Ouarzazate} über Hostelsclub.\\ 
    Reference & Book the best hostels in \textbf{Ouarzazate} with Hostelsclub.\\\hline
    \citealt{Transformer} & Book the best hostels in \textbf{ouarzazazate} with Hostelsclub.\\
    \citealt{D19-5611} & Book the best hostels in \textbf{Ouarzate} with Hostelsclub.\\
    \citealt{SyntaxInfusedNMT} & Book the best hostels in \textbf{ouarzazazate} with Hostelsclub.\\
    Aspect Augmented NMT& Book the best hostels in \textbf{Ouarzazate} with Hostelsclub.\\
    \hline\hline
    Source & Die \textbf{Deutsche Bahn} will im kommenden Jahr die Kinzigtal-Bahnstrecke verbessern.\\
    Reference & The \textbf{Deutsche Bahn} hopes to improve the Kinzigtal railway line in the coming year.\\\hline
    \citealt{Transformer} & \textbf{The German Railway} wants to improve the Kinzig valley railway line next year.\\
    \citealt{D19-5611} & \textbf{Christian Deutsche Bahn} intends to improve the Kinzig valley railway next year.\\
    \citealt{SyntaxInfusedNMT} & \textbf{The German Railway} wants to improve the kinziggia railway line next year.\\ 
    Aspect Augmented NMT & \textbf{Deutsche Bahn} wants to improve the Kinzig valley railway in the coming year.\\
    \hline\hline
  \end{tabular}
  }
  \caption{Examples of improved translation quality of WMT data where \textit{word-shape} and \textit{sub-word position} aspect vectors have helped the model choose a better sequence of sub-words when it faces out-of-vocabulary tokens.}
  \label{table:example2}
\end{table*} 

\begin{table*}[tbh!]
    \centering
    \addtolength{\tabcolsep}{1pt}
    \small{
     \begin{tabular}{lccccccc}
     \hline\hline
    \multirow{2}{*}{\textbf{a) M30k\textsuperscript{$\dagger$}}}& \multicolumn{2}{c}{\textbf{German to English}} & & & &\multicolumn{2}{c}{\textbf{German to French}} \\ \cline{2-3} \cline{7-8} 
    & \textbf{val} & \textbf{test2016} & ~ & ~ & & \textbf{val} & \textbf{test2016} \\ \hline
    \begin{tabular}[c]{@{}l@{}}\citealt{Transformer}\end{tabular}                         & 37.20 & 36.56 & ~ & ~ & ~ & 53.22 & 52.58 \\ 
    \begin{tabular}[c]{@{}l@{}}\citealt{SyntaxInfusedNMT}\end{tabular}                    & 38.14 & 37.13 & ~ & ~ & ~ & 54.18 & 54.37 \\ 
    \begin{tabular}[c]{@{}l@{}}\citealt{D19-5611}\\\ \ (bert freeze)\end{tabular}         & 38.44 & 37.42 & ~ & ~ & ~ & 55.10 & 54.50\\ 
    \begin{tabular}[c]{@{}l@{}}Aspect Augmented \\\ \ \  +M30k asp.\ vectors\end{tabular} & \textbf{39.22} & 38.17 & ~ & ~ & ~ & \textbf{56.21} & \textbf{56.40}\\ 
    \begin{tabular}[c]{@{}l@{}}Aspect Augmented \\\ \ \  +WMT asp.\ vectors\end{tabular}  & 38.90 & \textbf{38.57} & ~ & ~ & ~ & 56.12 & 55.98\\ 
    \hline\hline
    ~\\
    \hline\hline
    \textbf{b) IWSLT\textsuperscript{$\dagger$}}& \textbf{dev2010} & \textbf{tst2010} & \textbf{tst2011} & \textbf{tst2012} & \textbf{tst2013} & \textbf{tst2014} & \textbf{tst2015}\\ \hline
    \begin{tabular}[c]{@{}l@{}}\citealt{Transformer}\end{tabular}                            & 31.82 & 31.99 & 34.57 & 32.65 & 32.49 & 30.65 & 31.13 \\
    \begin{tabular}[c]{@{}l@{}}\citealt{SyntaxInfusedNMT}\end{tabular}                       & 32.91 & 32.95 & 35.35 & 33.10 & 33.17 & 31.32 & 31.90 \\
    \begin{tabular}[c]{@{}l@{}}\citealt{D19-5611}\\\ \ (bert freeze)\end{tabular}            & 33.34 & 32.78 & 35.42 & 33.12 & 33.20 & 31.22 & 31.45 \\
    \begin{tabular}[c]{@{}l@{}}Aspect Augmented \\\ \ \  +IWSLT asp.\ vectors\end{tabular}   & 32.86 & 32.86 & 35.38 & 33.43 & 33.23 & 31.37 & 31.87 \\
    \begin{tabular}[c]{@{}l@{}}Aspect Augmented \\\ \ \  +WMT asp.\ vectors\end{tabular}     & \textbf{33.78} & \textbf{33.56} & \textbf{36.14} & \textbf{33.51} & \textbf{33.98} & \textbf{31.86} & \textbf{32.37} \\
    \hline\hline
    ~\\
    \hline\hline
    \textbf{c) WMT\textsuperscript{$\dagger$}}& \textbf{wmt\_val} & \textbf{nt2014} & \textbf{nt2015} & \textbf{nt2016} & \textbf{nt2017} & \textbf{nt2018} & \textbf{nt2019}\\ \hline
    \begin{tabular}[c]{@{}l@{}}\citealt{Transformer}\end{tabular}                        & \textbf{30.65} & 33.80 & 33.70 & \textbf{37.10} & 34.44 & 37.81 & 36.05 \\ 
    \begin{tabular}[c]{@{}l@{}}\citealt{SyntaxInfusedNMT}\end{tabular}                   & 29.23 & 31.57 & 31.61 & 34.05 & 31.87 & 35.18 & 33.60 \\ 
    \begin{tabular}[c]{@{}l@{}}\citealt{D19-5611}\\\ \ (bert freeze)\end{tabular}        & 30.39 & 33.46 & 33.20 & 36.13 & 33.73 & 37.24 & 35.68 \\ 
    \begin{tabular}[c]{@{}l@{}}Aspect Augmented \\\ \ \  +WMT asp.\ vectors\end{tabular} & 30.61 & \textbf{33.97} & \textbf{33.99} & 37.01 & \textbf{34.71} & \textbf{38.17} & \textbf{36.48} \\ 
    \hline\hline
    \end{tabular}
    \caption{Evaluated \textsc{METEOR} score (calculated using the tool provided by Alon Lavie (\href{https://www.cs.cmu.edu/~alavie/METEOR/download/meteor-1.5.tar.gz}{https://www.cs.cmu.edu/ $\sim$alavie/METEOR/}; version 1.5)) results on M30k, IWSLT, and WMT datasets.\\\textsuperscript{$\dagger$} Each experiment was repeated three times, and we report the average in this table.\label{tab:meteor_results}}
    }
    \vspace{-0.3cm}
\end{table*}

Aside from performance, our model is approximately 5 times faster than syntax-infused translation model \citep{SyntaxInfusedNMT} while demanding less number of trainable parameters. Although it is not as fast as bert-freeze model \citep{D19-5611} in large settings (because of the size of computations required for calculating the linguistic embedding), it is comparable in speed to bert-freeze in medium and small scale settings. Appendix \ref{sec:apx_result_analysis} contains some additional insights regarding how aspect vectors can help translation systems trained on different dataset sizes.

Tables \ref{table:example1} and \ref{table:example2} demonstrate some examples of cases where aspect vectors has been useful in improving the translation quality.

\section{Conclusion and Future Work}
\vspace{-0.1cm}
In this paper, we proposed a simple method of extracting linguistic information from BERT contextual embeddings and integrating them into neural machine translation framework. We showed that the linguistic aspect vectors provide the translation models with out-of-domain knowledge which not only improves the translation quality but also helps the model to better deal with out-of-vocabulary words.
In the future, we would like to reconsider the integration module as a multi-head attention module, except that it will attend to different linguistic aspects of the current sub-word or sub-word tokens of a single word.
Increasing the number of linguistic aspects (especially the use of syntactic dependencies and morphology) and studying the effects of the aspect vector size on the quality of generated translations are other directions of future research. We would also like to examine the effectiveness of aspect vectors trained on large out-of-domain data in low-resource settings and explore the effects of using linguistic aspect vectors in tasks other than machine translation.

\section*{Acknowledgments}
\vspace{-0.1cm}
We would like to thank the anonymous reviewers for their helpful comments. The research was partially supported by the Natural Sciences and Engineering Research Council of Canada grants NSERC RGPIN-2018-06437 and RGPAS-2018-522574 and a Department of National Defence (DND) and NSERC grant DGDND-2018-00025.

\bibliography{references}
\bibliographystyle{acl_natbib}
\appendix

\section{Appendices}

\subsection{Implementation Details}\label{sec:baselines_}
In this section, we provide implementation details that could not be placed in the main write-up due to space limitations, but we believe are quite helpful for replication of our work. We divide this section into two parts, one focused on linguistic aspect vector extraction (Section \ref{sec:lave}) and the other on linguistic aspect integrated machine translation (Section \ref{sec:laimt}).

\subsubsection{Linguistic Aspect Vector Extraction Implementation Details}\label{sec:apx_aspect_details}
The pre-trained spaCy tagger that we used in our experiments is trained on the word-level while the pre-trained BERT operates on sub-word level\footnote{The alignment is non-trivial e.g. ``\texttt{hadn't}'' is tokenized to ``\texttt{hadn}'' and ``\texttt{'t}'' by spaCy and to ``\texttt{had}'' and ``\texttt{n't}'' by BERT, causing many-to-many alignments.}. The two sequences need to be aligned, so we can assign aspect attributes to BERT sub-word tokens. Inspired by \citet{J93-1004}, we align the two sequences using a heuristic divide-and-conquer monotonic alignment technique which finds the parts of the two sequences that are certainly equal and aligns the parts in between using recursive calls to itself\footnote{\href{https://github.com/sfu-natlang/SFUTranslate/blob/master/translate/readers/sequence\_alignment.py\#L51}{https://github.com/sfu-natlang/SFUTranslate/translate/}}.

Next, we explain how we implement the aspect extractors. We implement our aspect extractors using PyTorch framework and initialize them using Xavier initialization \citep{XavierInit}. We perform backpropagation using SGD (initial learning rate of 0.05, momentum value of 0.9, gradient clip norm of 5.0). 
To cope with inequality in the frequency of the different tags in each aspect tag set ($t_a$, see \S\ref{sec:aspect}), we practice weighted backpropagation with weights proportional to the inverse frequency of each tag. We decay learning rate with a factor of 0.9 when the loss value stops improving. 

\subsubsection{Linguistic Aspect Integrated Machine Translation Implementation Details}\label{sec:apx_nmt_details}
We implement our baseline transformer model using the guidelines suggested by \citet{W18-2509} in our translation toolkit SFUTranslate
\begin{table}[t]
    \centering
    {
    \begin{tabular}{llll}
        \hline\hline
        \textbf{Dataset} & \textbf{WMT} & \textbf{IWSLT} & \textbf{M30k}\\\hline
        N & 6 & 6 & 4\\
        $d_{model}$ & 512 & 256 & 256\\
        $d_{ff}$ & 2048 & 512 & 512\\
        h & 8 & 4 & 4\\
        opt factor & 1 & 2 & 1\\
        opt warmup & 4000 & 8000 & 2000\\
        grad accumulation & 8 & 2 & 1\\
        batch size\textsuperscript{$\ast$} & 4096 & 4096 & 2560\\
        epochs & 7 & 20 & 20\\
        \hline\hline
    \end{tabular}
    \caption{The transformer model settings for each dataset given the training data size. ``N'' is the number of layers in both encoder and decoder. Please see \S\ref{sec:nmt_and_bert} for more information about model parameters.\\
    \textsuperscript{\small{$\ast$}}The maximum number of sub-word tokens per batch.}
    
    \label{tab:model_settings_}
    }
    \vspace{-0.44cm}
    
\end{table}
and extend it for implementing the aspect-augmented model as well as the syntax-infused transformer and transformer with bert-freeze input setting. Table \ref{tab:model_settings_} provides the configuration settings for each of the models used in our experiments.

We use the pre-trained WordPiece\footnote{\href{https://github.com/huggingface/tokenizers}{https://github.com/huggingface/tokenizers}} \citep{WordPiece} tokenizer packaged and shipped with BERT (containing 31,102 sub-word tokens for German language) to tokenize the source side data, and tokenize the target side data with MosesTokenizer\footnote{\href{https://github.com/alvations/sacremoses}{https://github.com/alvations/sacremoses}} followed by the same WordPiece tokenizer model, trained on target data, to split the target tokens into sub-tokens. We set the target side WordPiece vocabulary size to 30,000 sub-words for English and French. Our models share the vocabulary
and embedding modules of both source and target \citep{E17-2025} since both source and target are trained in sub-word space. The shared vocabulary sizes of M30k (German to English), M30k (German to French), IWSLT, and WMT are 16645, 16074, 40807, 47940, respectively.

We generate target sentences using beam search with beam size 4 and length normalization factor \citep{GNMT} of 0.6. We merge the WordPiece tokens in the generated sentences (a post-processing step to create words) and use MosesDetokenizer\footnote{\href{https://github.com/alvations/sacremoses}{https://github.com/alvations/sacremoses}} to detokenize the generated outputs. We use Moses recaser\footnote{\href{https://github.com/moses-smt/mosesdecoder/tree/master/scripts/recaser}{https://github.com/moses-smt/mosesdecoder}} to produce cased translation outputs. We use \texttt{mteval-v14.pl} script for cased \textsc{Bleu} evaluation.

For all models, we set positional encoding max length to 4096, dropout to 0.1, loss prediction smoothing to 0.1, and initialize the models using Xavier initialization \citep{XavierInit}. We train all models using \texttt{NoamOpt} optimizer \citep{W18-2509} and perform the gradient accumulation trick \citep{W18-6301} with one update per a number of batches (Table \ref{tab:model_settings_}; \texttt{grad accumulation}
) to simulate larger batch sizes on a single GPU.

\vspace{-0.15cm}
\subsection{Additional Analysis of Linguistic Aspect Integrated Machine Translation Results}\label{sec:apx_result_analysis}

In this section, we analyze the results of our aspect integrated translation experiments. We provide our analysis in two parts, one for small and medium sized datasets and the other for large ones.

For smaller datasets (containing a few hundred thousand sentence pairs or less), the broader perspective of BERT knowledge is helpful in limiting the search space for the model. So using our technique, the translation model receives more information regarding the general use cases of (locally) rare words. Linguistic aspect vectors also help the model better understand less familiar (in comparison to what is frequent in its limited size training data) syntactic structures in input sentences. This is why we believe aspect vectors can be helpful in low-resource settings.

Improving models with large amounts of data (with several million sentence pairs) is a challenging task. 
The best practice in training neural translation models is to initialize the embedding module with small random values and let the model search through the parameter space to find the optimal parameter settings.
Extracted aspect vectors, as an external source of monolingual knowledge on the source side, are a more reasonable starting point for large models than random initialization. Integrating aspect vectors thus helps these models find a better path towards the optimal point(s) and increases the chances of the model ending up in a more desirable point in search space.

\end{document}